\documentclass[letterpaper, 10 pt, conference]{ieeeconf}  

\usepackage{graphicx}
\usepackage{epsfig,subfigure}
\usepackage{amssymb,amsmath}
\usepackage{enumerate}
\usepackage{latexsym}
\usepackage{booktabs}
\usepackage{multirow} 
\usepackage{array} 
\usepackage{cite}
\usepackage{comment}
\usepackage{todonotes}
\usepackage{authblk}
\usepackage[algoruled]{algorithm2e}
\usepackage{url}
\usepackage{epstopdf}
\usepackage{placeins}

\IEEEoverridecommandlockouts                              

\title{\LARGE \bf
Synthetic Image Augmentation for Improved Classification using Generative Adversarial Networks
}

\author{Keval Doshi
\thanks{*This work was not supported by any organization}
\thanks{$^{1}$Keval Doshi is with Department of Electrical Engineering, University of South Florida, Tampa, USA.
        {\tt\small kevaldoshi@mail.usf.edu}}%
}

\begin{document}

\maketitle
\thispagestyle{empty}
\pagestyle{empty}

\begin{abstract}

Object detection and recognition has been an ongoing research topic for a long time in the field of computer vision. Even in robotics, detecting the state of an object by a robot still remains a challenging task. Also, collecting data for each possible state is also not feasible. In this literature, we use a deep convolutional neural network with SVM as a classifier to help with recognizing the state of a cooking object. We also study how a generative adversarial network can be used for synthetic data augmentation and improving the classification accuracy. The main motivation behind this work is to estimate how well a robot could recognize the current state of an object.

\end{abstract}

\section{INTRODUCTION}

In the past few years, there has been significant development in the field of computer vision especially in object recognition. There have been some very effective
and popular methods which have been developed such as Support
Vector Machines, Convolutional Neural Networks, Artificial Immune System, etc. Among those, Deep learning has been gaining a lot of popularity as it has proved to be one of the most effective techniques for high dimensional datasets such as images. In deep learning, a convolutional neural network (CNN, or ConvNet) is a class of deep neural networks, which use a variation of multilayer perceptrons designed to require minimal preprocessing. Typical CNNs used for object recognition use dozens of layers which builds thousands of parameters in the network. The key algorithms of CNN can be traced back to the late 1980s. CNNs saw heavy use in the 1990s. It however fell out of fashion with the rise of support vector machines. Interest in CNNs was rekindled again in 2012 when it proved to drastically reduce the error rates and improve classification rates on popular dataset such as MNIST \cite{deng2012mnist}, ImageNet \cite{deng2009imagenet}, etc.  However, object state recognition has not been addressed as much even though it has many important applications in robotics, medical, gaming and many more.  

One such important application is to teach a robot how to cook, which however involves several important steps one of which is recognizing the state of food items. For example, a robot should be able to recognize the object (e.g. an onion) and identify the state (e.g. diced onion). In robotics, objects at different states would require different manipulation strategies. A significant difficulty with image recognition tasks is collecting generous amount of training data. For example, collecting hundreds of training images for different states of each object might not even be feasible. In this case, synthetic data augmentation needs to be considered. In this paper we analyze a small section of the state  recognition  problem,  which  is  classification. Classification  mainly  helps  to  tell  what  sate  an  object  is  in. We consider Inception v3 as our base model to extract features which are then classified using a Support Vector Machine (SVM) \cite{cortes1995support} to solve a eleven class image recognition problem. Also, generative adversarial networks (GANs) \cite{goodfellow2014generative} have been used for synthetic data augmentation. Specifically, we use a CycleGAN \cite{zhu2017unpaired} for unpaired image-to-image translation. The final model is shown to achieve 81\% accuracy. The primary objective of this research is to develop an automated recognition system which:
\newline
\begin{enumerate}
\vspace{-2mm}
\item Classifies cooking items according to its present state.
\vspace{1mm}
\item Is robust to challenging real-world conditions such as varying video resolution and lighting conditions. 
\vspace{2mm}
\end{enumerate}
The outline of the current study is as follows: We look at some of the related works in Section II. In Section III, a brief description of the dataset along with different data augmentation methods used is provided. The fourth section consists of an overview of the proposed approach to classification is provided. This section will highlight the computer vision algorithms selected for the purposes of this study. The training and fine-tuning of the algorithm will also be discussed. In the fifth section, the final results along with a discussion of the proposed model as compared to different models is provided. Lastly, concluding remarks, recommendations, and additional research needs for future
154 are presented in the sixth section.

\section{Related Works}

A major breakthrough in using neural networks was made by Yann Le Cun when he proposed an algorithm to train Neural Networks. The innovation \cite{lecun1990handwritten} was to simplify the architecture and to use the back-propagation algorithm to train the entire system. The approach was successful in performing tasks such as OCR and handwriting recognition. Recently, a hierarchical state space markov model was used to recognize cooking activities using Object based task grouping (OTG) in \cite{lade2010task}. In robotics, knowing the object states and recognizing archiving the desired states are very important. Sun \cite{sun2014object,sun2015modeling,jelodar2018identifying} presented a novel object learning approach for robots to understand the object’s interaction from human. Particularly, a bayesian network was used to represent the knowledge learned based on which the recognition reliability of objects and human was improved, which would further be extrapolated to help the robots to properly understand a pair of objects. Huang \cite{sun2016robotic} focused on the requirements of grasps from the physical interactions in instrument manipulation. In \cite{schlafly2019feature}, a SVM model was used for feature selection in gait classification of leg length and distal mass. A discussion on synthetic data augmentation using generative adversarial networks for improved liver classification is provided in \cite{frid2018synthetic}. 
\section{Dataset and Preprocessing}

The data used in this paper includes eleven different classes: creamy paste, diced, floured, grated, juiced, julienne, mixed, other, peeled, sliced and whole with 17 cooking objects(carrot, tomato, pepper, onions, cheese,
etc). In our work, dataset version 1.2 of the state recognition challenge from USF RPAL lab was used. The dataset was first annotated to classify an image to one of the eleven classes. It consists of 9309 images out of which 6348 images were used as training data and 1377 images were used as validation data wheras the rest was reserved for testing. However, 6348 images are not sufficient for such a intricate classification task. Hence, several augmentation methods were used to further supplement the dataset such as height shifting, rescaling, rotation as well as vertical and horizontal flipping. The precise augmentation factors used are presented in Table I.

\begin{table}[]
\centering
\caption{Parameters used for data augmentation}
\label{Table}
\begin{tabular}{|l|l|}
\hline
\textbf{Name}       & \textbf{Augmentation Factors} \\ \hline
Rotation            & 45                            \\ \hline
Shear               & 0.2                           \\ \hline
Zoom                & 0.2                           \\ \hline
Rescale             & 1/255                         \\ \hline
Horizontal Flipping & True                          \\ \hline
Vertical Flipping   & True                          \\ \hline
\end{tabular}
\end{table}

Also, a new promising approach for training a model that synthesizes synthetic images known as Generative Adversarial Networks (GANs) is used. GANs have gained great popularity in the computer vision community and different variations of GANs were recently proposed for generating high quality realistic natural images \cite{radford2015unsupervised,odena2017conditional}. Also, there has been an increase in the applications which have applied the GAN framework \cite{costa2017towards,schlegl2017unsupervised,nie2017medical}. Most studies have employed the image-to-image GAN technique to create label-to-segmentation translation, segmentation to-image translation or medical cross modality translations. Some studies have been inspired by the GAN method for image in painting. In the current study we investigate the applicability of GAN framework to synthesize new training images. In particular, we use a Cycle GAN which is capable of translating an image from domain $X$ to to a target domain $Y$ in the absence of paired examples. The objective is to learn a mapping $G: X \rightarrow Y$ such that the distribution of images from $G(X)$ is indistinguishable from the distribution $Y$ using an adversarial loss. Because this mapping is highly under-constrained, it is coupled with an inverse mapping $F: Y \rightarrow X$ and a cycle consistency loss is introduced to push $F(G(X)) ≈ X$ (and vice versa) \cite{zhu2017unpaired}.

\begin{figure}[!htb]
\centering
\includegraphics[width=0.45\textwidth]{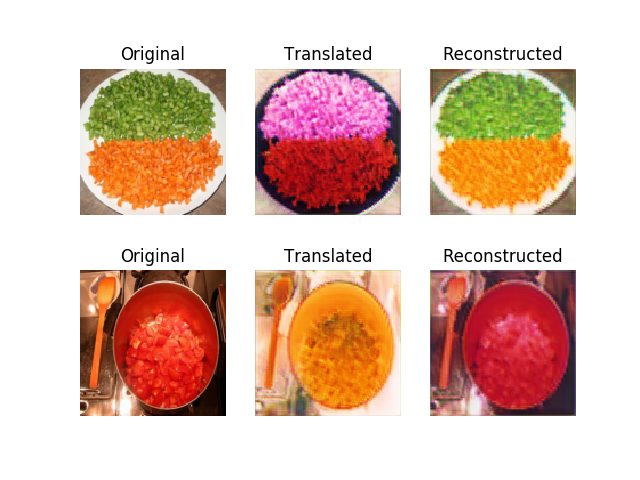}
\caption{Synthetic Data Augmentation using Cycle GAN.}
\label{f:SDA}
\end{figure}

\begin{figure}[!htb]
\centering
\includegraphics[width=0.45\textwidth]{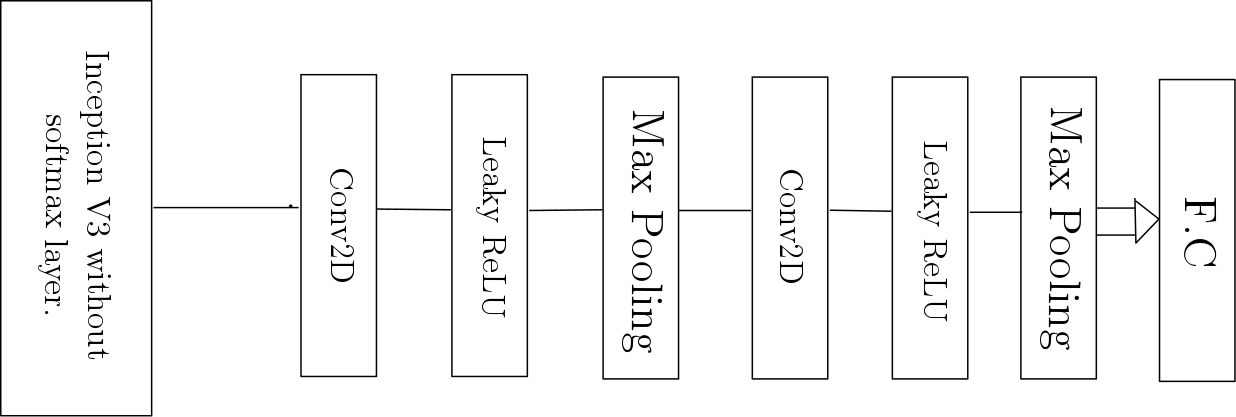}
\caption{Modified architecture for Deep Convolutional Neural Network.}
\label{f:modified}
\end{figure}
\section{Methodology}

The classification system developed in this study consists of a CNN which extracts unique feature descriptors after warping the images to fixed square size of (299 x 299). Then, the feature descriptor for each test image is classified through a linear SVM. The following section explains in detail the procedure followed to build the CNN classifier.

Convolutional Neural Networks are computationally very expensive to train which earlier made them very unpopular for practical applications. However, with the recent rise in efficient graphical processing units along with deep learning libraries such as Tensorflow \cite{abadi2016tensorflow}, Torch etc have made CNNs a viable option. In this study, a dual RTX 2070 GPU system was used for training the model. Model training involved two main steps: Base model selection and classifier specific fine tuning.

\textbf{Base model selection:} In real world examples, it might not always be feasible to train a network from scratch. In such cases, we can use a technique known as transfer learning in which a pre-trained model is re-purposed on a second related task. The pre-trained model used in this work is Inception v3, which is a deep convolutional neural network architecture based on GoogLeNet and developed by Google. This architecture achieved the highest accuracy for classification and detection in the ImageNet \cite{deng2009imagenet} Large-Scale Visual Recognition Challenge 2014 (ILSVRC14) \cite{berg2010large}. Inception performs the utilization of the computational resources in an improved way with an emphasis on depth and width while maintaining a lower computational complexity. 

\textbf{Modified CNN architecture:} As shown in Fig. \ref{f:modified}, the base pre-trained model has been modified to better suit our application. First, the last layer of the pre-trained model is removed and a convolution layer is added. A convolution layer generates output in form of tensors as a convolution kernel convoluted with the input of this layer. Then, a batch normalization layer is added, which provides a way to shift inputs closer to zero
or mean of all values in the input array. The primary advantage of this technique is to help in faster learning and getting higher accuracy. In a deep convolution neural network, as the
data flows further, the weights and parameters adjust their values. This flow can make the intermediate data too big for computation or too small to give correct prediction. However, this problem can be avoided by normalizing the data in mini-batches. Next an activation layer is added with Leaky Rectified Linear Unit (Leaky ReLU) as the activation function. The same layers are added once more before adding a dropout layer, which is used for regularization. Regularization helps the model to generalize better by reducing the risk of over-fitting. There is a risk of over-fitting if the size of data set is too small as compared to the number
of parameters needed to be learned. A dropout layer randomly removes some nodes and their connections in the network. The last layer is a fully connected layer.

\textbf{Specific fine tuning:} To adapt the pre-trained model to the proposed task (state recognition), the CNN model parameters are fine-tuned. First, the last softmax classification layer of the pre-trained model is replaced with eleven classes. Stochastic gradient descent is used with a learning rate of 0.001, which allows fine-tuning to make progress while not clobbering the initialization. While training the model, first the entire pre-trained model is frozen and only the last newly added layers are trained for 70 epochs. Then, the first 5 layers of the pre-trained model are fine tuned for 40 epochs to generate the final model. The features extracted are then used to train a SVM model using MATLAB machine learning toolbox.

In the next section, we take a closer look at how different models perform and the accuracy achieved by the best model.

\section{Evaluation and Results}

\begin{figure}[!htb]
\centering
\includegraphics[width=0.45\textwidth]{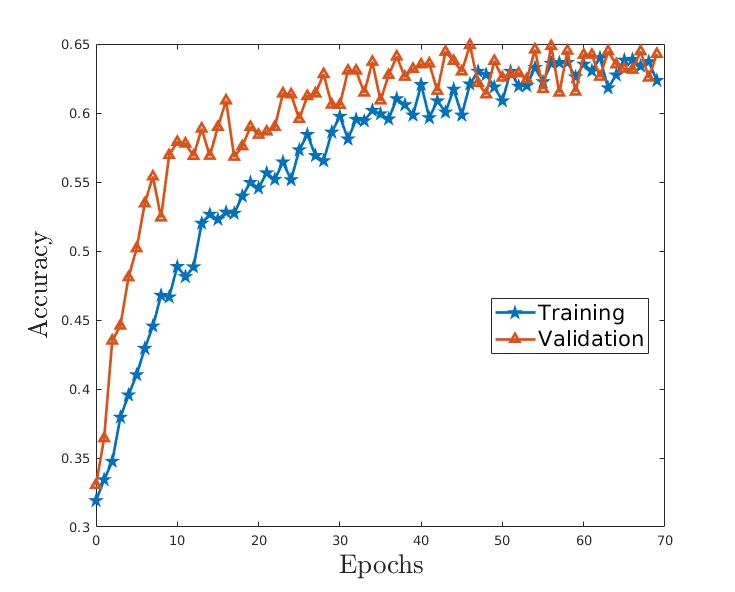}
\caption{Accuracy vs Epochs for training and validation during training the newly added layers.}
\label{f:Byz4}
\end{figure}

\begin{figure}[!htb]
\centering
\includegraphics[width=0.45\textwidth]{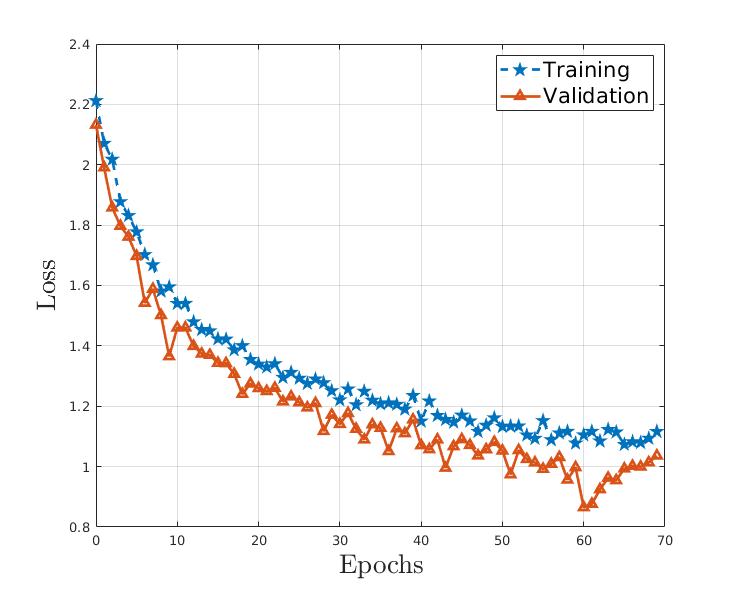}
\caption{Loss vs Epochs for training and validation during training the newly added layers.}
\label{f:Byz3}
\end{figure}

\begin{figure}[!htb]
\centering
\includegraphics[width=0.45\textwidth]{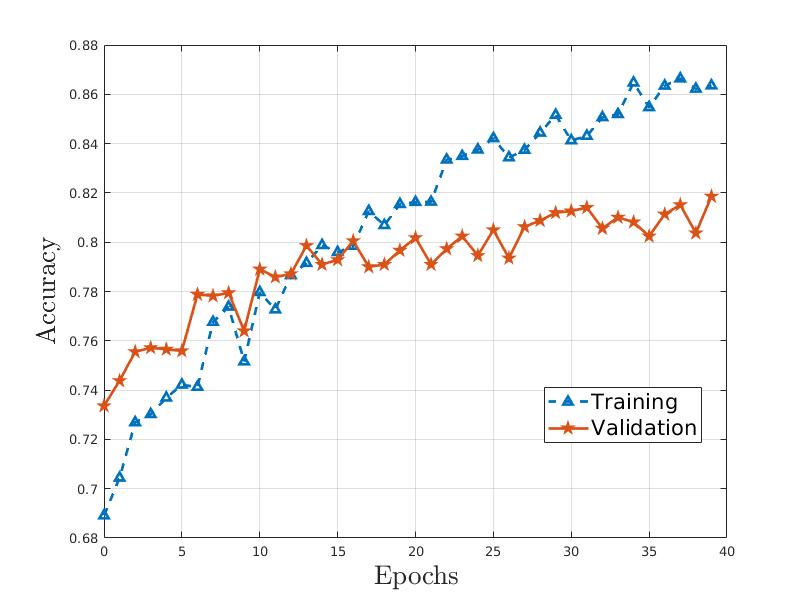}
\caption{Accuracy vs Epochs for training and validation during fine-tuning the model.}
\label{f:Byz1}
\end{figure}

\begin{figure}[!htb]
\centering
\includegraphics[width=0.45\textwidth]{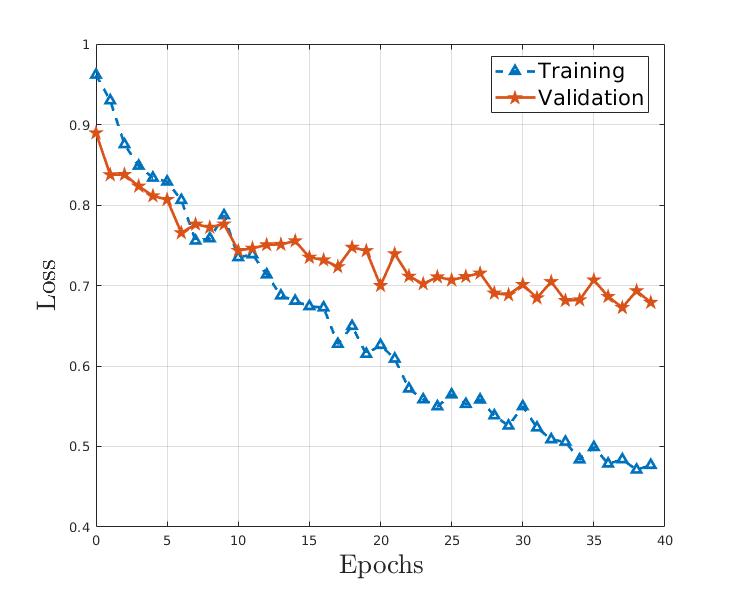}
\caption{Loss vs Epochs for training and validation during fine-tuning the model.}
\label{f:Byz2}
\end{figure}

As shown in Table II, we performed various experiments to generate the best model. From these experiments, it can be determined that adding more layers does infact help in improving the results. This can be attributed to an increase in the number of parameters because of which the CNN can learn better. The best result was obtained by fine tuning an additional set of layers of the pretrained network, which was Inception V3 in this case. Fine tuning is was done on top of an already trained model, which was the best model considering all previous training examples. The training and validation accuracies and losses are shown in Fig. \ref{f:Byz4},  \ref{f:Byz3}, \ref{f:Byz1}, \ref{f:Byz2}  respectively.  However, to further improve the accuracy, SVM with a linear kernel was used. A comparision of different kernels for SVM is provided in Table II. In Table III, we provide a comparision between the performance of different models. The best performance is when two convolutional layers are added along fine-tuning and using SVM as a classifier whereas the worst performance is by Vanilla Inception v3 which does not have any additional layers or fine tuning.     

\begin{table}[]
\centering
\centering
\caption{Performance of different SVM kernels}
\label{Table III}
\begin{tabular}{|l|l|}
\hline
SVM Kernel & Accuracy \\ \hline
Linear     & 81.3\%   \\ \hline
Quadratic  & 79.8\%   \\ \hline
RBF        & 79\%     \\ \hline
\end{tabular}
\end{table}

We see that linear SVM performs better as compared to SVM with Quadratic or RBF kernel.

\begin{table}[]
\centering
\caption{Performance of different models on the testing data}
\label{my-label}

\begin{tabular}{|l|l|l|l|}
\hline
\multicolumn{1}{|c|}{\begin{tabular}[c]{@{}c@{}}Model \\ Trained\end{tabular}}                                                          & \multicolumn{1}{c|}{\begin{tabular}[c]{@{}c@{}}Validation\\ Accuracy\end{tabular}} & \multicolumn{1}{c|}{\begin{tabular}[c]{@{}c@{}}Validation\\ Loss\end{tabular}} & \multicolumn{1}{c|}{Epochs} \\ \hline
Vanilla Inception v3                                                                                                                    & 67\%                                                                               & 1.21                                                                           & 70                         \\ \hline
\begin{tabular}[c]{@{}l@{}}Inception V3 with one \\ convolutional layer and\\  max pooling.\end{tabular}                                & 71\%                                                                               & 1.02                                                                           & 70                         \\ \hline
\begin{tabular}[c]{@{}l@{}}Inception V3 with two \\ convolutional layers and\\  max pooling.\end{tabular}                               & 72\%                                                                               & 0.97                                                                           & 70                         \\ \hline
\begin{tabular}[c]{@{}l@{}}Inception V3 with two \\ convolutional layers and\\  max pooling along with \\ fine tuning.\end{tabular}     & 76\%                                                                               & 0.88                                                                           & 70 + 40                    \\ \hline
\begin{tabular}[c]{@{}l@{}}Inception V3 with two \\ convolutional layers, \\ max pooling along with\\ fine tuning and SVM.\end{tabular} & 81\%                                                                               & 0.72                                                                           & 70 + 40                    \\ \hline
\end{tabular}%
\end{table}

\section{Conclusion}

In this study we propose a model for detecting the current state of an object. The cooking objects have eleven different states and the best classification accuracy achieved on the test data was 81\%. Before this work, there were several studies for object detection but detecting the state of an object has not been researched as much. However, there are a few shortcomings that still need to be addressed for this experiment to viable in a real setting. The training data needs to be properly cleaned as some of the images are have multiple states or are unclear. These images could in turn result in generating a less accurate model. Also, using GAN as way of synthetic data augmentation needs to be explored as a Cycle GAN is only able to output images with size 128x128 with a majority of them being slightly blurry. To implement this in a real life situation a significant amount of study is still required.

\bibliographystyle{unsrt}
\bibliography{main.bib}

\addtolength{\textheight}{-12cm}   




\end{document}